# Bayesian Networks for Dependability Analysis: an Application to Digital Control Reliability

**Luigi Portinale and Andrea Bobbio**
Dipartimento di Scienze e Tecnologie Avanzate
Università del Piemonte Orientale "A. Avogadro" - Alessandria (ITALY)

## Abstract

Bayesian Networks (*BN*) provide robust probabilistic methods of reasoning under uncertainty, but despite their formal grounds are strictly based on the notion of conditional dependence, not much attention has been paid so far to their use in *dependability analysis*. The aim of this paper is to propose *BN* as a suitable tool for dependability analysis, by challenging the formalism with basic issues arising in dependability tasks. We will discuss how both modeling and analysis issues can be naturally dealt with by *BN*. Moreover, we will show how some limitations intrinsic to combinatorial dependability methods such as Fault Trees can be overcome using *BN*. This will be pursued through the study of a real-world example concerning the reliability analysis of a redundant digital Programmable Logic Controller (PLC) with majority voting 2:3

## 1 Introduction

Dependability analysis involves a set of methodologies dealing with the reliability aspects of large, safety-critical systems [9, 10]. Two main categories of approaches can be identified: *combinatorial methods* and *state-space based methods*. The first category requires a description of the system to be analyzed in terms of components and their interactions. In particular, components are modeled as binary events corresponding to *component up* and *component down* respectively, while interactions are usually described by means of boolean AND/OR gates. The advantage of such techniques is that they are component-oriented and that simple formalisms like *Fault Trees* (*FT*) can be adopted. However, some major simplificative assumptions are usually needed, both at the modeling and analysis level.

On the contrary, state-space approaches rely on the specification of the whole set of possible states of the system and on the modeling of the possible transitions among them. Markov Chain methods are usually adopted in this case, giving the analyst the possibility of explicitly considering all possible interactions among components. The main disadvantage is that, even for not very large systems, using directly the approach may be unfeasable due to the huge dimensionality of the considered state-space. The usual solutions is then to resort to partial state descriptions like *Petri Nets* [11] or Bayesian Networks based models [5]. However, state-space approaches are essentially used when emphasis is on temporal evolution of the systems, while combinatorial methods concentrate more on static aspects as the logical interactions of the components[1].

In this paper we aim at considering combinatorial methods for dependability analysis, by showing how basic problems addressed by the approach can be profitably dealt with by using Bayesian Networks. In particular we will concentrate on *Fault Tree Analysis* (*FTA*) and on the modeling and analysis issues that are involved in this task. We will show that any *FT* can be mapped into a *BN* and that any analysis that can be performed on the *FT* can be performed by means of *BN* inference. In addition, standard modeling technique of *BN* like noisy gates, multi-state variables or common cause identification, can overcome some basic modeling limitations of fault-trees, while interesting analysis neglected in *FTA* can be naturally obtained by classical *BN* inference.

The paper is organized as follows: section 2 discusses basics of fault-tree analysis, while in section 3 an algorithm for mapping a *FT* into a *BN* is proposed; in section 4 the suitability of a *BN* for dependability analysis is investigated and in section 5 basic extensions to

---
[1] As we will see, this does not means that time is not considered, but only that temporal evolution is not directly modeled.

*FTA* provided by Bayesian networks are discussed, together with the improvements they provide both at the modeling and at the analysis level.

## 2   Fault-Tree Analysis

Combinatorial models of dependability have a poor modeling power coupled with a high analytical tractability based on the assumption of statistically independent component. Among them, *Fault-Tree Analysis (FTA)* has become very popular for the analysis of large safety-critical systems [9, 10]. The goal of *FTA* is to represent the combination of elementary causes, called the *primary events*, that lead to the occurrence of an undesired catastrophic event, the so called *Top Event (TE)*. *FTA* is carried out in two steps: a qualitative step, in which the list of all the possible combinations of primary events (the *minimal cut-sets*) that give rise to the *TE* is determined and a quantitative step where, if probability values can be assigned (directly or by computation) to all the events appearing in the tree, the probability of occurrence of the *TE* can also be calculated.

The qualitative analysis step is considered very important in dependability analysis and safety studies, since it allows the analyst to enumerate all the possible causes of failure for the system and to rank them according to a very simple severity measure given by the (prior) failure probabilities of components. However, because of the simplicity of the model, only prior failure probability can be considered and no kind of scenarios more specific than cut-sets can be computed.

The major weak point of the *FTA* methodology (as well as of any combinatorial technique), is the fact that the events must be considered as statistically independent. The aim of the present paper is to show that, even by representing knowledge as in combinatorial models, alternative formalisms like Bayesian networks can be profitably adopted to overcome their limitations. Let us now introduce some basic notions concerning *FTA*:

**Definition 1** *A fault tree (FT) is a tree comprising five types of nodes:* primary events, events, *two types of elementary logical gates, namely* AND *gates and* OR *gates, and one derived logical gate, namely the k out of n gate k : n (see Figure 1).*

*The root of a FT is an event called the* Top Event *(TE), representing the failure of the whole system. Each event has exactly one successor which can be either a primary event or a logical gate. A logical gate has two or more successors, which are all events; a k out of n gate has exactly n successor events. The leaves of the tree are all primary events.*

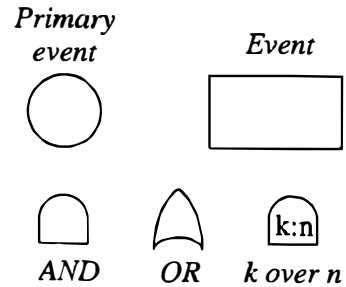

Figure 1: Basic Notation for *FTA*

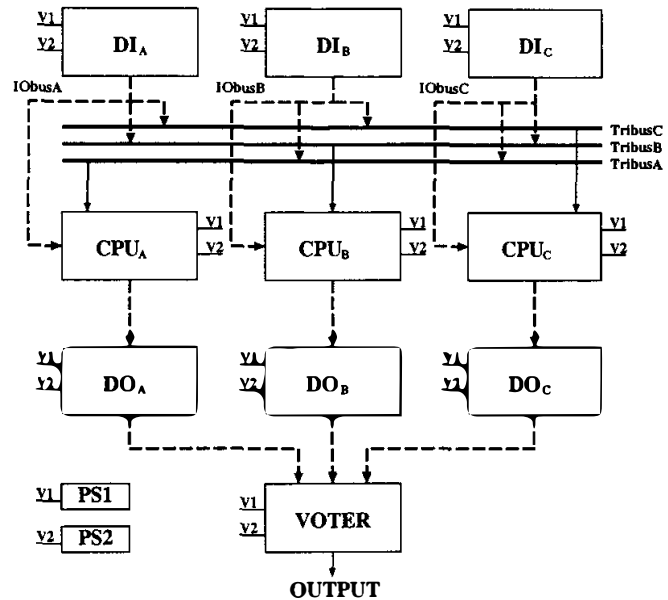

Figure 2: The PLC Case Study

The events considered in any *FT* are binary events, that is they may just assume logical values *true* or *false*. Since events are used to represent the functional behavior of a given sub-system, the logical values are put in correspondence with *faulty* and *working* behavior respectively. In particular, primary events are associated to system components. A *FT* is then constructed by combining primary and general events with logical gates, in order to model the functional behavior of the whole system to be analyzed.

**Example 1.** Let use introduce a simple but real PLC system, that we will use as a case study throughout the paper. The block diagram of such a system is shown in figure 2. The PLC system is intended to process a digital signal by means of suitable processing units; in particular, a redundancy technique is adopted, in order to achieve fault tolerance, so three different *channels* are used to process the signal and a *voter hardware device*, with majority voting 2 : 3, is collecting channel results to produce the output. For each channel (identified as channels $Ch_A, Ch_B$ and $Ch_C$ respectively) a *digital input unit* $(DI)$, a *processing unit* $(CPU)$ and a *digital output unit* $(DO)$ are employed. The digital signal elaborated by a given channel is transmitted among units through a special bus called $IObus$. Moreover, redundancy is present also at the $CPU$ level; indeed, each $CPU$ receives from its digital input unit the signal to be elaborated, but it also receives a copy of the signal from other input channels. For doing this, a bunch of three buses is used, called $Tribus_A$, $Tribus_B$ and $Tribus_C$ respectively; the $IObus_X$ of channel $Ch_X$ delivers the signal of the digital input of channel $Ch_X$ $(DI_X)$ to tri-buses of other channels (i.e. to $Tribus_Y$ with $Y \neq X$) and $CPU_X$ reads the signal from other input channels from $Tribus_X$. Each $CPU$ performs then a software majority voting 2 : 3 for determining the input signal. Finally, the system is completed by a redundancy on the power supply system: two independent *power supply units* $(PS_1$ and $PS_2)$ are connected to other components, in such a way that the simultaneous failure of both $PS$ units is needed to prevent the system to work.

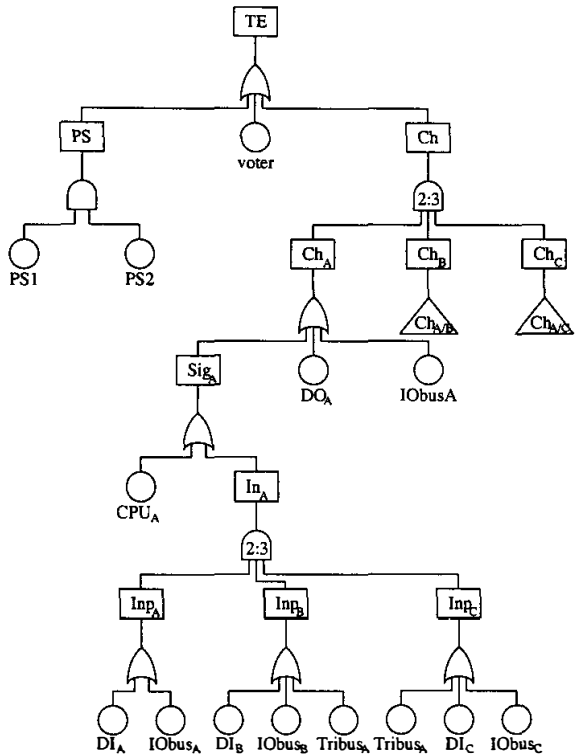

Figure 3: Fault Tree for the PLC System

The $FT$ for this PLC system is reported in figure 3. Notice that 2 : 3 gates are used to model the fault of the input part of each channel (since each $CPU$ adopts a 2 : 3 majority voting) and the fault corresponding to the fault of at least 2 channels (since the voter also uses a 2 : 3 majority voting). Notice also that sub-trees rooted at a given channel $X$ are similarly replicated for other channel different from $X$; indeed, notation $Ch_{A/B}$ and $Ch_{A/C}$ is used to indicate that sub-trees rooted at $Ch_B$ and $Ch_C$ are equal to the sub-tree rooted at $Ch_A$ with $A$ substituted with $B$ and $C$ respectively[2].

Given a model like the $FT$ of figure 3, some typical analysis can be performed.

**Qualitative Analysis.** Given the $FT$, to determine the *minimal cut-sets* $(MCS)$, i.e. the minimal (with respect to set inclusion) set of primary events causing the occurrence of the $TE$(the *prime implicants* of the $TE$).

**Quantitative Analysis.** *i)* Given probabilistic information about truth of primary events (i.e. about failure of components), to determine the probability of occurrence of any event in the $FT$ and in particular of the $TE$. *ii)* Given $MCS$, to determine their importance, also called their *unreliability*.

$MCS$ determination is usually based on minimization techniques on the set of boolean functions represented by the gates of the $FT$. The cardinality of a cut-set is called its *order*. The $FT$ of figure 3 has 59 $MCS$, one of order 1 (corresponding to the voter faulty) and the remaining 58 of order 2. For instance, a $MCS$ of order 2 that can immediately be derived from the $FT$ of figure 3 is $\{PS_1, PS_2\}$ corresponding to the simultaneous failure of both power suppliers.

Concerning quantitative analysis, the most common assumption made in $FTA$ is to assume that components corresponding to primary events have an exponentially distributed failure time. This means that the probability of having component $C$ faulty at time $t$ (alternatively the probability of occurrence of the primary event $C = faulty$) is $P(C = faulty, t) = 1 - e^{-\lambda_C t}$, where $\lambda_C$ is the *failure rate* of component $C$.

Given the failure rates of each component, $FTA$ can determine at any time instant $t$, the probability of occurrence of any event and in particular the probability of system failure (the $TE$) at time $t$. Moreover, the so-called unreliability of $MCS$ at time $t$ can be computed. This corresponds to the probability of the joint occurrence of events in the cut-set at time $t$; because of the independence assumption made on the occurrence of primary events, this corresponds to the product of the probability of occurrence of each primary event in the cut-set at time $t^3$.

---
[2] Actually events $Inp_A$, $Inp_B$ and $Inp_C$ have to be dealt appropriately by making a double indexing with respect to the channels; this should be more clear in the following.

[3] Other kind of quantitative information such as the *mean time to failure* and the *variance of time to failure*

| Component | Failure Rate (f/h) | Failure Prob. |
|---|---|---|
| IObus | $\lambda_{IO} = 2.0 \ 10^{-9}$ | 0.00080 |
| Tribus | $\lambda_{Tri} = 2.0 \ 10^{-9}$ | 0.00080 |
| Voter | $\lambda_V = 6.6 \ 10^{-8}$ | 0.02605 |
| DO | $\lambda_{DO} = 2.45 \ 10^{-7}$ | 0.09335 |
| DI | $\lambda_{DI} = 2.8 \ 10^{-7}$ | 0.10595 |
| PS | $\lambda_{PS} = 3.37 \ 10^{-7}$ | 0.12611 |
| CPU | $\lambda_{CPU} = 4.82 \ 10^{-7}$ | 0.17535 |

Table 1: Failure Rates and Probabilities for PLC Components

| MCS | Unrel. | Post. Unrel. | Post. Prob. |
|---|---|---|---|
| $\{CPU_A, CPU_B\}$ | 0.03075 | 0.13943 | 0.04533 |
| $\{CPU_B, CPU_C\}$ | 0.03075 | 0.13943 | 0.04533 |
| $\{CPU_A, CPU_C\}$ | 0.03075 | 0.13943 | 0.04533 |
| $\{Voter\}$ | 0.02605 | 0.11812 | 0.02681 |
| $\{CPU_A, DO_C\}$ | 0.01637 | 0.07423 | 0.02195 |
| $\{CPU_A, DO_B\}$ | 0.01637 | 0.07423 | 0.02195 |
| $\{CPU_B, DO_A\}$ | 0.01637 | 0.07423 | 0.02195 |
| $\{CPU_B, DO_C\}$ | 0.01637 | 0.07423 | 0.02195 |
| $\{CPU_C, DO_A\}$ | 0.01637 | 0.07423 | 0.02195 |
| $\{CPU_C, DO_B\}$ | 0.01637 | 0.07423 | 0.02195 |
| $\{PS_1, PS_2\}$ | 0.01590 | 0.07212 | 0.02088 |

Table 2: Top 11 MCS for PLC System

**Example 2.** In the system of figure 2, the failure rates (in terms of $failure/hour$) shown in table 1 can be assumed by considering exponentially distributed failure time. Table 1 also shows the failure probability of components after $4 \cdot 10^5$ hours of system operation. FTA can then compute the probability of system failure at time $t = 4 \cdot 10^5$ hours as $P(TE) = 0.22053$. Similarly, the probability of any other event of the FT can be obtained. For example the probability of having a failure in the input part of a channel $(P(In_X = faulty) = 0.03248, X = A, B, C)$ or the probability of failure of at least two channels $(P(Ch = faulty) = 0.18674)$.

Finally, MCS can be ranked in order of unreliability; table 2 shows the first 11 MCS and in the second column (Unrel.) their corresponding unreliability. We can easily verify that the most critical components are the CPUs, since the cut-sets involving a failure of at least two CPUs are those showing larger unreliability.

In the next sections we will show that, by using Bayesian networks, besides to perform the above analyses we can augment both the modeling and the analysis power in dependability tasks.

## 3   Mapping Fault Trees to Bayesian Networks

Given a FT, it is straightforward to map it into a binary BN, where every variable has two admissible values: *false* corresponding to a *normal* or *working* value and *true* corresponding to a *faulty* or *not-working* value. The conversion can be obtained as follows:

- for each *leaf node* of the FT, create a *root node* in the BN; however, if several leaves of the FT represent the same primary event (i.e. the same component), create just one root node in the BN to represent all of them.

- for each pair *(gate, output-event)* of the FT, create a corresponding *node* in the BN;

- connects nodes in the BN as corresponding nodes are connected in the FT;

- for each node of the BN created from an AND (respectively OR) gate, create a *Conditional Probability Table* (CPT) such that the node is true with probability 1 iff all parent nodes are true (respectively iff at least one parent node is true);

- for each node created from a *k out of n* gate, create a CPT such that the node is true with probability 1 iff at least *k* out of *n* parent nodes are true.

Prior probabilities on root nodes have to be established by considering a given time point $t$. Given a root node $C$, the probability of $C = true$ will be set to the probability of occurrence of the corresponding primary event at time $t$ (i.e. $P(C = faulty), t) = 1 - e^{-\lambda_C t}$). It should be clear that from the above conversion non-root nodes of the BN are actually *deterministic nodes*, i.e. special *chance (random) nodes* with associated a deterministic function for their value determination [8].

Figure 4 shows the structure of the BN for the PLC system of figure 2 and derived from the FT of figure 3. Random variable nodes (i.e. root nodes) are shown as gray ovals, while deterministic variable nodes are empty ovals. Notice that, as mentioned in section 2, events of type $Inp$ actually need two indices to distinguish the cases where a Tribus is involved from those where it is not. Prior probabilities for roots can be obtained from table 1 for time $t = 4 \cdot 10^5$ hours. Near each deterministic node is indicated the boolean function represented. The user has just to specify the type of function: CPT specification can then be automatically obtained[4].

---

can be obtained, but this is usually a matter of general reliability analysis not peculiar to FTA.

[4] Alternatively, the number of required probabilities can be reduced by techniques like the *asymmetric assessment*

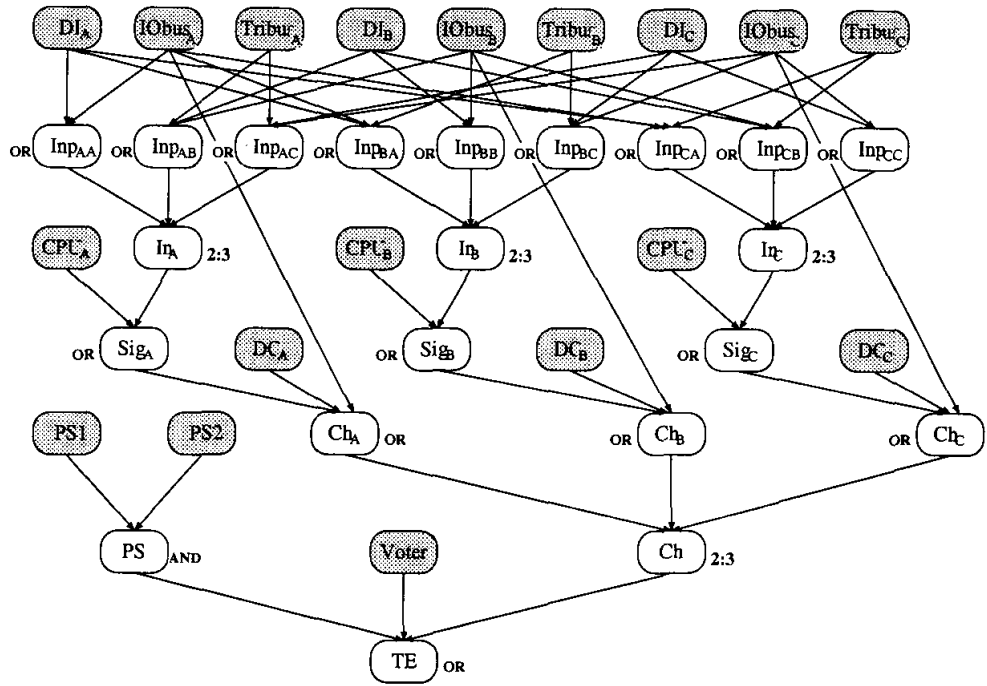

Figure 4: Bayesian Net for the PLC System: first letter of a double index denotes a channel

## 4 Dependability Analysis with Bayesian Networks

Typical *FTA* can be performed quite naturally in a *BN* setting. In particular, concerning *MCS* determination (i.e. qualitative analysis), because any deterministic node of the *BN* correspond to a boolean gate of the *FT*, any kind of technique adopted in *FTA* for computing *MCS* can be applied also for the corresponding $BN^5$. However, as we shall see in the following, the *BN* framework can refine the notion of cut-set when uncertainty about combinatorial knowledge has to be introduced (see section 5). Concerning quantitative analysis, a *BN* can in principle computes the posterior probability of any set of variables $Q$ given the evidence set $E$ (i.e. $P(Q|E)$), so any kind of probabilistic computation that can be performed in *FTA* can also be performed by *BN* inference. In particular the following parameters can be computed at time $t$:

**probability of occurrence of an event**; it corresponds to the computation of the prior marginal at time $t$ on the node corresponding to the given event;

**unreliability of *MCS***; it corresponds to the computation of the joint prior on nodes mentioned in the cut-set at time $t^6$.

Concerning the first point, the marginals of all the failure events represented in the net can be computed by *BN* probability propagation [12]. However, standard *BN* inference usually deals with posterior probability computation, while the above issue are only related to prior information (i.e. to prediction of particular events at time $t$). In fact, by considering the occurrence of a failure, posterior information can be very relevant for dependability and reliability aspects.

**Example 3.** Suppose the system failure has been observed in the PLC system (i.e. the *TE* has occurred). A *Belief Updating* [12] algorithm can be used to compute marginal posteriors on root nodes. Results are summarized in table 3. Notice for instance that, differently from the prior information, after observing the fault, a digital output unit is less reliable than a digital input one; this is reasonable since the difference in prior reliability was very small and the input part of a channel has more redundancy that the output part. This information cannot be directly obtained in *FTA*; it can be deduced by the fact that cut-sets involving digital output units have a higher unreliability than those containing digital inputs (see table 2). Marginal posterior probability of nodes corresponding to non primary events (i.e. to sub-systems failure) can similarly be computed.

---

[6].

[5]More specific techniques relying on the "logical semantics" of a *BN* in terms of Horn clauses [13, 14] can also be devised, by adopting abductive reasoning.

[6]As in *FTA* this is obtained by multiplying the prior of the true value of each node in the cut-set.

| Component | Post. Failure Prob. |
|---|---|
| Tribus | 0.00175 |
| IObus | 0.00208 |
| Voter | 0.11812 |
| DI | 0.17167 |
| PS | 0.17603 |
| DO | 0.20433 |
| CPU | 0.38382 |

Table 3: Posterior Probabilities for PLC Components

Regarding *MCS* analysis, *BN* inference can be more precise than usual *FTA*; indeed, *(Composite) Belief Revision* [12] algorithms can be used to compute posterior probability of *MCS*. Even if the ranking provided by prior unreliability of *MCS* is the same as the one provided by posterior unreliability [7], posterior unreliability is a more reliable measure. Table 2 reports such parameters on the third column. Notice that, even if such values are in principle computable also with *FTA* (after the computation of the probability of the *TE*), *FTA* tools usually report just the information of table 2.

Finally, composite queries can provide results more specific than *MCS*; indeed, in a *MCS* no commitments is made to unmentioned events. More specific information can be obtained by setting for query in *BN* inference all primary events. If we consider a component-oriented framework for system analysis (as is done in reliability), then *MCS* corresponds to *partial* or *kernel diagnoses*, while a composite query on every primary event will produce *complete diagnoses* (in terms of component behavioral modes) [3]. While in a logical setting kernel diagnoses can be suitably adopted for concisely explain a system failure, in probabilistic analysis this is no longer true, since marginalization operations can provide counter-intuitive results [12]. If we interpret each entry of the first column of table 2 as assigning the faulty mode to mentioned components and the working mode to unmentioned ones, we can interpret the *MCS* as diagnoses whose posterior probability if given in the last column of the table. Notice that, even if the top 11 diagnoses of our problem actually correspond to top 11 *MCS* in the same order of probability, this is not true in general, since diagnoses corresponding to non-minimal cut-sets may have a larger probability than some diagnoses corresponding to *MCS*. For instance, even if the PLC system has 59 *MCS*, the 18*th* (in order of probability) diagnosis is the one corresponding to all CPUs faulty and every other component working (probability 0.00963) and this does not correspond to a *MCS*.

## 5 Augmenting Modeling Power

There are essentially three main features of *BN* formalism that may be exploited to improve combinatorial dependability analysis: *noisy gates*, *multi-state variables* and *sequential failure dependence*.

### 5.1 Noisy gates

Differently from fault-trees, in a *BN* dependency relations between events or variables are not restricted to be deterministic. In dependability terms, this corresponds to being able to model uncertainty in the behavior of the "gates" that in a *FT* represent interactions between sub-systems. Of particular attention for reliability aspects is one peculiar modeling feature often used in building *BN* models: *noisy gates*. The most common kind of noisy gate is the *noisy-or* model [12] and its generalizations [7]; this kind of model can be profitably used in dependability, since it allows a simple probabilistic generalization of boolean gates.

Consider, in the PLC case study, a situation in which more fault-tolerance is added to the system by means of a third spare power supplier that is available under certain circumstances; in particular, it is shared by other systems and it is available only if it is not already in use. In this situation, it is no longer true that when both modeled suppliers are down, also the system is certainly down, since the control system could switch to the spare supplier. Suppose that, from statistical information, we know that the spare supplier is available the 30% of time when the *PS* sub-system is down: we can model the uncertainty about the availability of the spare supplier by transforming the gate corresponding to the *TE* in a noisy-or gate with the following parameters (remember that *TE= true* means system failure):

$c(TE = true | PS = working, Voter = working, Ch = working) = 0$

$c(TE = true | PS = faulty) = 0.7$

$c(TE = true | Voter = faulty) = P(TE = true | Ch = faulty) = 1$

As done in [2] we use $c$ instead of $P$ to emphasize the fact that these are not standard conditional probabilities. The noisy-or independence assumption about causes of the *TE* is in this case reasonable and we could conclude that the system is down when only the *PS* sub-system is down with probability

$P(TE = true | PS = faulty, Voter = working, Ch =$

---

[7]Since *MCS* are prime implicants of the *TE*, the conditional probability of *TE* given a *MCS* is 1, so the posterior of a *MCS* given the *TE* differs from the prior only for the constant $P(TE)^{-1}$.

$working) = 1 - (1 - 0.7) = 0.7$

corresponding to the percentage of times that the spare supplier is not available, while if all causes are present we obtain

$P(TE = true|PS = faulty, Voter = faulty, Ch = faulty) = 1 - (0.3 \cdot 0 \cdot 0) = 1$

Moreover, another important modeling issue in dependability analysis is the problem usually referred as the *common causes* problem, where the system may go down, even in the presence of components up. This usually refer to the presence of some common unknown cause of failure that has been neglected in the model. This is naturally treated in a *BN* by means of *leak probabilities* [15]. In the above example, we may have that the $TE$ may be true even when both $PS$, $Voter$ and $Ch$ sub-systems are functioning, because of a common cause problem; by assuming a leak probability $l = 1 \cdot 10^{-4}$ the first parameter becomes

$c(TE = true|PS = working, Voter = working, Ch = working) = l = 0.0001$

and the unreliability of the system when sub-systems $PS$, $Voter$ and $Ch$ are down becomes

$P(TE = true|PS = faulty, Voter = working, Ch = working) = 1 - ((1 - 0.7) \cdot (1 - 0.0001)) = 0.70003$

which is slightly larger than in the case where no common cause problem was present.

Dually from noisy-or, we could also use a *noisy-and* gate to generalize AND gates of the *FT*. For instance, by considering the possibility of failure of wire connections from power supply units, we can model the fact that, even if only one supplier is down, the $PS$ sub-system is also down (because connections from the other supplier are not working). More specifically, if we assume that this event has a probability of 0.01, then we can specify the noisy-and as

$c(PS = faulty|PS1 = faulty, PS2 = faulty) = 1$

$c(PS = faulty|PS1 = working) = 0.01$

$c(PS = faulty|PS2 = working) = 0.01$

and then to compute

$P(PS = faulty|PS1 = working, PS2 = working) = 0.01 \cdot 0.01 = 0.0001$

corresponding to the simultaneous independent failure of connections from both supplier.

### 5.2 Multi-state variables

The working/faulty dichotomy of *FTA* can be improved towards the most reasonable approach of dealing with variables having more than two values (multi-state variables). This is particularly useful for primary events, since in this way the system component they refer to can be modeled by means of multiple behavioral modes [4]. In fact, components may manifest more than one failure mode (e.g open/short) and the failure modes may have a very different effect on the system operation (e.g. fail-safe/fail-danger). Suppose to consider a three-state component whose states are identified as *working (w)*, *fail-open (f-o)* and *fail-short (f-s)*. In *FTA* the component failure modes must be modeled as two independent binary events ($w/f-o$) and ($w/f-s$); however, to make the model correct, a (non-standard) *XOR* gate must be inserted between *f-o* and *f-s* since they are mutually exclusive events. On the contrary, Bayesian networks can include *n*-ary variables by adjusting the entries of the *CPT*. Saving in assessment provided by classical noisy-or in case of binary variables can now be obtained through generalizations to *n*-ary variables, like for instance the *noisy-max* gate [15]. In the next subsection we will return more specifically on this point.

### 5.3 Sequentially Dependent Failures

Another modeling issue that may be quite problematic to deal with by using fault-trees is the problem of components failing in some dependent way. Bayesian networks may address this point by making explicit such a dependency in the structure of the net. Consider for instance the case in which, in the PLC of figure 2, power suppliers may induce a CPU failure when failing; this can be due to the fact that a possible fault for power supplier is an over-voltage, causing a possible CPU damage (and then a CPU fault). While it is not possible to model this kind of information in a *FT*, in a *BN* it can be naturally modeled by connecting each power supplier to the CPU nodes. In particular, one can even be more precise, by resorting to multi-state variable modeling; indeed, each $PS_i$ unit ($i = 1, 2$) can be modeled with three ordered values, *working, over_voltage, dead*, and connected as a parent to each CPU node $CPU_X$ ($X = \{A, B, C\}$) with the following noisy-max parameters:

$c(CPU_X = faulty|PS_1 = working, PS_2 = working) = 0$

$c(CPU_X = faulty|PS_i = over\_voltage) = 0.66667$

$c(CPU_X = faulty|PS_i = dead) = 1$

The probability of getting a CPU fault when both the suppliers are in over-voltage can then be computed from noisy-max model as

$P(CPU_X = faulty|PS_1 = over\_voltage, PS_2 = over\_voltage) = 1 - (0.33333 \cdot 0.33333) = 0.88889$

This shows how a flexible combination of basic features of a *BN* can naturally overcome basic limitations of *FTA*.

Finally, it is worth noting that, if uncertainty about combinatorial knowledge on components has to be modeled (either through noisy gates simplifications or through complete *CPT* specifications), the *FTA* notion of *MCS* does no longer make sense. In this case, the computation of diagnoses intended as composite beliefs on primary events as shown in section 4, provides a natural counterpart of the *MCS* concept.

## 6   Conclusions

In the present paper we have discussed the suitability of Bayesian networks for classical combinatorial dependability analysis. A preliminary analysis on this topic was proposed in [1] (where no formal comparison with classical dependability analysis was provided) and in [16] (where the starting point are reliability block diagrams). We have shown that combinatorial formalisms like fault-trees can be formally mapped into binary *BN* and that classical *FTA* can be naturally performed through *BN* inference. Basic features of *BN* allow to overcome limitations of *FTA* both at the modeling and at the analysis level. Among those, the possibility of modeling uncertainty at gates level, the use of multi-state variables and the modeling of dependent failures have been investigated, as well as analytical improvements over *FTA* like general posterior probabilistic inference. This has been done by considering a real-world dependability case study concerning a digital PLC, where the use of a *BN* methodology has shown the importance of all the above mentioned improvements.

**Acknowledgements**

The research presented in this paper has been partially funded by ENEA, the Italian Board for Energy and Environment, which provided the PLC case study. We would like to thank B. D'Ambrosio for having made available the SPI system and for all the answers to our questions.